\pdfoutput=1

\documentclass[11pt]{article}

\usepackage[]{emnlp2021}

\usepackage{times}
\usepackage{latexsym}

\usepackage[T1]{fontenc}

\usepackage[utf8]{inputenc}

\usepackage{microtype}

\usepackage{soul}
\usepackage{url}
\usepackage{graphicx}
\usepackage{amsmath}
\usepackage{booktabs}
\usepackage{subfigure}
\usepackage{amsmath, amssymb}
\usepackage{makecell}
\usepackage{multirow}
\usepackage{footnote}
\usepackage{amssymb}
\usepackage{nameref}

\title{Unsupervised Open-Domain Question Answering}

\author{Pengfei Zhu \\
  Shanghai JiaoTong \\
  University \\
  \texttt{zhupf97@sjtu} \\
  \texttt{.edu.cn} \\\And
  Xiaoguang Li \\
  Huawei Noah's\\
  Ark Lab \\
  \texttt{lixiaoguang11@}\\
  \texttt{huawei.com} \\\And
  Jian Li \\
  Huawei Noah's\\
  Ark Lab \\
  \texttt{lijian703@}\\
  \texttt{huawei.com}\\\And
  Hai Zhao \\
  Shanghai JiaoTong \\
  University \\
  \texttt{zhaohai@cs.sjtu}\\
  \texttt{.edu.cn} \\}

\begin{document}
\maketitle
\begin{abstract}
Open-domain Question Answering (ODQA) has achieved significant results in terms of supervised learning manner. However, data annotation cannot also be irresistible for its huge demand in an open domain. Though unsupervised QA or unsupervised Machine Reading Comprehension (MRC) has been tried more or less, unsupervised ODQA has not been touched according to our best knowledge. This paper thus pioneers the work of unsupervised ODQA by formally introducing the task and proposing a series of key data construction methods. Our exploration in this work inspiringly shows unsupervised ODQA can reach up to 86\% performance of supervised ones.
\end{abstract}

\section{Introduction}
Open-domain Question Answering (ODQA) is the task of answering questions based on information from a very large collection of documents which has a variety of topics \cite{chen2020open}.
Unlike Machine Reading Comprehension (MRC) task where a passage containing evidences and answers is provided for each question, ODQA is more challenging as there is no such supporting passage beforehand.
ODQA systems need to go through a large collection of passages such as the whole Wikipedia to find the correct answer.

While tremendous progress on ODQA have been made based on pretrained language models such as BERT \cite{devlin-etal-2019-bert}, ELECTRA \cite{clark2020electric}, and T5 \cite{2020t5}, fine-tuning these language models requires large-scale labeled data, i.e., passage-question-answer triples \cite{lewis2019unsupervised}.
Apparently, it is costly and practically infeasible to manually create a dataset for every new domain.

Though previous studies which have made attempts in unsupervised MRC like \cite{lewis2019unsupervised,li2020harvesting,fabbri2020template,hong2020handling,perez-etal-2020-unsupervised}, as to our best knowledge, no such manner of attempts have been made in terms of ODQA. Thus in this paper, for the first time, we tackle the ODQA setting without human-annotated data, which we term Unsupervised ODQA (UODQA).
Concretely, our setting is: starting from an automatically generated question or question-like sentence, we employ a lexical-based retriever like BM25 to retrieve positive passages that contain the answer and negative passages without the answer, from the Wikipedia corpus.
Together with these, we can effectively train a question answering model which can handle multiple passages.

Unlike UQA which the supporting passage is certain for each question, UODQA needs to construct more than one passages through retrieval-based method and solve a multi-passage MRC problem.


As the first attempt to tackle UODQA, we propose a series of methods about how to synthesize data from a set of selected natural sentences and compare end-to-end performance, and finally we achieve up to 86\% performance of previous SOTA supervised method on three ODQA benchmarks.

\section{Related Work}

\subsection{Open-Domain Question Answering}
Open-domain Question Answering (ODQA) needs to find answers from tremendous open domain information such as Wikipedia or web pages. Traditional methods usually adopt \textbf{retriever-reader} architecture \cite{karpukhin2020dense}, which is to first retrieve relevant documents and then generate answers based on these retrieved documents, which is the main focus of our paper. Besides, there is also end-to-end method \cite{guu2020realm}, but it costs too much computation resources to be widely applied. The improvements of \textbf{retriever} \cite{izacard2020distilling} and \textbf{reader} \cite{izacard2020leveraging} are both critical for the overall performance, and there is still huge room for improvements.

\subsection{Unsupervised Question Answering}
Unsupervised Question Answering (UQA) is to alleviate the problem of huge cost of data annotation. Generally speaking, the key issue of UQA aims at automatically generating context-question-answer triples from publicly available data. \cite{lewis2019unsupervised} uses an Unsupervised Neural Machine Translation method to generate questions. \cite{fabbri2020template} proposes to retrieve relevant sentence that contains the answer and reform the sentence with template-based rules to generate questions.  \cite{li2020harvesting} proposes an iterative process to refine the generated questions turn by turn. \cite{hong2020handling} proposes paraphrasing and trimming methods to respectively solve the problem of word overlap and unanswerable generated questions.

\section{Task Definition}

For UODQA task, there is no limitation to use or construct data for training, only development and test sets from ODQA benchmark have to be used for evaluation and fair comparison. Therefore we will focus on data construction hereafter.

Based on a specific corpus $C$, several $<Q,P^+,P^-,A>$ triples are constructed, For each constructed example, $Q$ denotes the question, $A$ denotes the answer, $P^+$ denotes multiple positive passages that contains the answer supporting the question to solve, $P^-$ denotes multiple negative passages that do not contain the answer, and help make the model learn to distinguish distracting information. To train a reader model, these data are leveraged to learn a function $F(Q,P^+,P^-)=A$.

\section{Method}

\begin{figure*}[!htb]
	\centering
	\includegraphics[width=1.0 \textwidth]{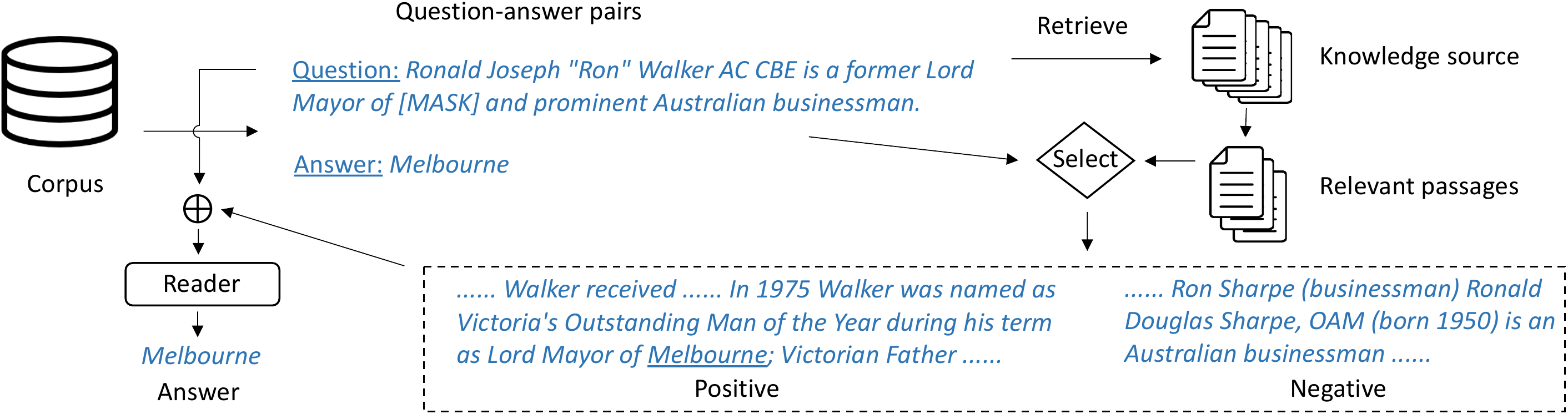}
	\caption{Our proposed method for synthesizing data and training.}
	\label{fig:architecture}
\end{figure*}

\subsection{Data Construction}
\label{sec:data_construct}
The procedure is shown in Figure \ref{fig:architecture}. The purpose is to automatically construct $<Q,P^+,P^-,A>$ triples for model training. Obviously, the quality of constructed data decides whether a model can be trained well. 

Firstly, based on some specific corpus $C$, we select a set of sentences to construct $<Q,A>$. In ODQA, most of the questions are factoid. Many works show that knowing Named Entities (NEs) may help construct $<Q,A>$ pairs \cite{glass-etal-2020-span,guu2020realm}, thus a good practice is to select NEs as $A$ in the constructed data. Meanwhile the sentence where the NE is from is $Q$ after the selected NE is masked. The constructed $Q$ is thus a pseudo-question, or conceptually defined as \textbf{Information Request}. 
Note that there is an obvious expression difference between real questions and our constructed information request, in which the former usually starts with an interrogative and the latter does not so but is just plain statement. Despite of this syntactic difference, both questions may relate to the same concerned factoids and be used for effective model training.
Meanwhile, to align the train data and test data, many works aim to do question generation based on constructed pseudo-questions, which is to reform statements as the expression of real questions. However, this procedure also introduces noises.

When selecting sentences to generate $Q$ from corpus $C$, previous works on UQA do not set constraints \cite{lewis2019unsupervised,li2020harvesting, hong2020handling}, which brings no guarantee the constructed information request is reasonable or answerable. Such none-guarantee will become much more severe in UODQA.
Basically, the source sentences selected need to have complete information. For example, \textit{``It was instead produced by Norro Wilson, although the album still had a distinguishable country pop sound.''} is ambiguous because of too many coreferences. Moreover, when selecting phrases as $A$, it needs to be answerable based on the constructed information request. For example, in sentence \textit{``Yao Ming played for the Houston Rockets of the National Basketball Association (NBA).''} if the phrase \textit{``Yao Ming''} is selected as $A$, the constructed \textit{``\underline{$\;\;\;\;\;\;$} played for the Houston Rockets of the National Basketball Association (NBA).''} is not certain and answerable.

To obtain $<Q,A>$ pairs with higher quality, we use sentences from the dataset in \cite{elsahar2019t}, which is an alignment corpus for WikiData and natural language. Each sentence is aligned with a Subject-Predicate-Object triple, and we select the object as $A$.

To obtain $P^+$ and $P^-$, our model retrieves documents from knowledge source $C^*$, and selects the documents containing $A$ as positive $P^+$ otherwise negative. This heuristic can not assure enough evidences but still make the model learn reasoning. To filter the trivial cases of $P^+$ that the context text surrounding the answer has too much overlap with that in the $Q$ so the answer can be simply generated based on shortcuts, we set a window size $n$ and check the left and right $n$-gram of the selected $A$. Thus, $<Q,P^+,P^-,A>$ triples are constructed.

\subsection{Model Training}
\label{sec:model_train}

Following previous common practice in ODQA, we adopt retriever-reader architecture to perform UODQA. BM25 serves as retrieval metric in an unsupervised manner. After retrieving top $K$ passages, a reader receives the questions and passages as input to output an answer. Following \cite{izacard2020leveraging}, we adopt a generative reader based on T5 \cite{2020t5}.

\section{Experiments and Analysis}
\begin{table}[]
\renewcommand\arraystretch{0.8}
\begin{tabular}{l|l|l|l}
\hline
Dataset           & \,\,train  & \,\,dev   & \,\,\,test   \\
\hline
\hline
Natural Questions & 79,168 & 8,757 & \,\,\,3,610  \\
WebQuestions      & \,\,\,3,417  & \,\,\,\;361   & \,\,\,2,032  \\
TriviaQA          & 78,785 & 8,837 & 11,313 \\
\hline
\end{tabular}
\caption{Data statistics of three datasets.}
\label{tab:data_statistics}
\end{table}

\subsection{Evaluation Settings}
The evaluation metrics are Exact Match (EM). For EM, if generated answer hits any one of the labeled list of possible golden answer, the sample is positive. The accuracy of EM is calculated as $EM = N^+/N$ where $N^+$ is number of positive samples and $N$ is number of all evaluated samples.

We evaluate our model on three ODQA benchmarks, Natural Questions \cite{kwiatkowski2019natural}, WebQuestions \cite{berant2013semantic} and TriviaQA \cite{joshi2017triviaqa}. Statistics are shown in Table \ref{tab:data_statistics}. The train/dev/test split follows \cite{lee-etal-2019-latent}. As this is an unsupervised ODQA task, we discard training set and only adopt development and test sets for evaluation.
\textbf{Natural Questions(NQ)} is commonly used ODQA benchmark which was constructed according to real Google search engine queries. The answers are short phrases from Wikipedia articles containing various NEs. 
\textbf{WebQuestions(WQ)} contains questions collected from Google Suggest API, and the answers are all entities from the structured knowledge base (Freebase).
\textbf{TriviaQA(TQA)} consists of trivia questions from online collection.

\subsection{Implementation Details}
We adopt dataset from \cite{elsahar2019t} and select sentences that have only one object to construct question-answer pairs. Sentences containing character number more than 250 or less than 50 are discarded. The object is used as answer and we use the token [MASK] to replace the answer in the sentence as the question. Following \cite{karpukhin2020dense}, the version of Wikipedia corpus we use is Dec. 20, 2018 dump and we split the whole corpus into 100-word segments as units of retrieval. For retrieving documents, we use Apache Lucene \footnote{https://lucene.apache.org/} to build index and perform BM25 retrieval. To filter $P^+$ using $n$-gram, we use $n$ as 3. We first retrieve top 100 documents, and select the top 40 documents to construct the input for reader. If none of top 40 documents contains the answer, we further find top 41-100 documents that contains the answer, and replace the 40th document with it, otherwise this sample is discarded. Finally, we obtain 844,100 samples to train for 2-3 day using 8 V100s.

We implement the reader following \cite{izacard2020leveraging} and perform training using learning rate of 1e-4, batch size of 256 and the number of concatenated passages each sample is 40. The model size setting we use is T5-base. We save and evaluate the model checkpoint every 500 training steps and stop training if the performance does not increase any more in 5 evaluations, and the checkpoint of best EM score is selected. 

\subsection{Results}
\begin{table}[]
\renewcommand\arraystretch{0.8}
\begin{tabular}{ll|p{0.8cm}|p{0.8cm}|p{0.8cm}}
\hline
                                                         &              & WQ & NQ & TQA \\ 
                                                         \hline 
                                                         \hline
\multicolumn{1}{p{1cm}|}{\multirow{2}{*}{sup.}} &   DPR\shortcite{karpukhin2020dense}        &    42.4               &  41.5            &    57.9      \\ 
\multicolumn{1}{p{1cm}|}{}                                    &   FiD\shortcite{izacard2020leveraging}        &     -              &      \textbf{51.4}        &   \textbf{67.6}       \\ \hline
\multicolumn{1}{p{1cm}|}{\multirow{5}{*}{unsup.}}         & RandSent$_{10}$     & 12.01             &      15.90        &      40.39    \\ 
\multicolumn{1}{p{1cm}|}{}                                    & RandEnt$_{10}$      &   15.01                &   18.14           &   45.38       \\ 
\multicolumn{1}{p{1cm}|}{}                                    & QuesGen$_{10}$      &   10.43                &  13.88            &  43.44        \\ 
\multicolumn{1}{p{1cm}|}{}                                    & OurMethod$_{10}$ &  16.14                &  18.73           &   46.64       \\ 
\multicolumn{1}{p{1cm}|}{}                                    & OurMethod$_{50}$ &  \textbf{18.60}                &    \textbf{20.69}          &    \textbf{50.23}      \\ \hline
\end{tabular}
\caption{Experimental results EM on test set of three datasets. ``sup.'' denotes supervised methods and ``unsup.'' denotes unsupervised methods. The subscript denotes number of passages to input the reader. }
\label{tab:results}
\end{table}

As shown in Table \ref{tab:results}, we perform experiments based on four kinds of settings, to study to what extent the quality of constructed training data affects performance. \textit{RandSent} means we select random sentences from Wikipedia articles and NEs to construct question-answer pairs. \textit{RandEnt} means we use data from \cite{elsahar2019t} and select a random NE from each sentence as answer. This expands the scope of types of answers and makes the model learn more diversified knowledge. \textit{QuesGen} means we perform a question generation step after obtaining the constructed data based on our method. This makes the expression of the pseudo-question more close to the real question and makes the model learn a question answering manner better, but it may hurts the reasonibility of constructed questions because of the noise introduced by question generation methods. Some examples are shown in Table \ref{tab:setting_example}.

\begin{table}[]
\renewcommand\arraystretch{0.9}
\centering
\begin{tabular}{l}
\hline
 Setting // Question // Answer \\
\hline
\hline
     RandSent   //       He had 16 caps for Italy, from \\ 1995 to [MASK], scoring 5 tries, 25 points in \\ aggregate. // 1999 \\
\hline
RandEnt  // [MASK] stiphra is a species of sea \\ snail, a marine gastropod mollusk in the family \\ Raphitomidae. // Daphnella    \\
 \hline
  QuesGen     // What is a multi-state state high-\\way in the New England region of the United \\ States, running across the southern parts of \\ New Hampshire, Vermont and Maine, and \\ numbered, owned, and maintained by each \\ of those states? // Route 9 \\
\hline
OurMethod // Ronald Joseph ``Ron'' Walker \\ AC CBE is a former Lord Mayor of [MASK] \\ and prominent Australian businessman. // Mel-\\bourne \\
\hline
\end{tabular}
\caption{Examples of the settings for comparison experiments.}
\label{tab:setting_example}
\end{table}

As shown in Table \ref{tab:results}, improving the quality of constructed training data improves the performance by a large margin. Moreover, the performance gap between supervised and unsupervised method indicates that the task is very challenging and shows huge space for improvements.

\subsection{Analysis and Discussion}

There are three main factors of the differene among different settings, reasonability, answerability and strategy to select answer span. Reasonability indicates to what extent the question conforms to the expression of natural language, answerability means whether the sentence describes an fact with accurate meanings and has enough information for deducing answer, and strategy to select answer span determines what knowledge the model learns.

For the setting of \textit{RandSent}, because random sentences are usually ambiguous and lack enough evidence to infer corresponding answer, the answerability is very weak. For the setting of \textit{RandEnt}, though the original sentence contains complete information and expresses accurate fact, the randomly masked NE may be too difficult to deduce. Compared with this, our strategy that only selects the object as answer is better, because in the structure of Subject-Predicate-Object, the object usually can be accurately deduced. \textit{QuesGen} attempts to reform the expression of question to make it more like a real question, however, it also introduces noise to do harm to the performance. For the purpose of implementing unsupervised manner, we only adopt simple rule-based question-generation method, which applies semantic role labeling on the original question and selects one of the parsed argument as answer, and converts the order and tense of the sentence to reform it as a question expression. It indicates that if the question generation method introduces too much noise and hurts the reasonability of sentences too much, it is even worse than doing nothing and maintaining the statement expression of original constructed information request sentences.

\section{Conclusion}
In this paper, we first propose the task of Unsupervised Open-domain Question Answering, and explore to what extent it can perform based on our suggested data construction methods. We compare several strategies for synthesizing better data, as a result achieve up to 86\% performance of previous supervised method. We hope this work inspires a new line of ODQA in the future and helps build more practical readers for real use.

\bibliography{anthology,custom}
\bibliographystyle{acl_natbib}

\end{document}